\documentclass[conference]{IEEEtran}

\IEEEoverridecommandlockouts
\usepackage{cite}
\usepackage{amsmath,amssymb,amsfonts}
\usepackage{algorithmic}
\usepackage{graphicx}
\usepackage{textcomp}
\usepackage{xcolor}
\def\BibTeX{{\rm B\kern-.05em{\sc i\kern-.025em b}\kern-.08em
    T\kern-.1667em\lower.7ex\hbox{E}\kern-.125emX}}

\usepackage{hyperref}

\usepackage{pgfplots}
\usepackage{pgf-pie}
\usepackage{caption}

\ifCLASSOPTIONcompsoc
    \usepackage[caption=false, font=normalsize, labelfont=sf, textfont=sf]{subfig}
\else
\usepackage[caption=false, font=footnotesize]{subfig}
\fi

\begin{document}

\title{Automated Quality Check of Sensor Data Annotations\\
}

\author{
\IEEEauthorblockN{Niklas Freund}
\IEEEauthorblockA{\textit{Digitale Schiene Deutschland} \\
\textit{DB InfraGO AG}\\
Berlin, Germany \\
niklas.freund@deutschebahn.com}
\and
\IEEEauthorblockN{Zekiye Ilknur-Öz}
\IEEEauthorblockA{\textit{Digitale Schiene Deutschland} \\
\textit{DB InfraGO AG}\\
Berlin, Germany \\
zekiye.ilknur-oez@deutschebahn.com}
\and
\IEEEauthorblockN{ Tobias Klockau}
\IEEEauthorblockA{\textit{Digitale Schiene Deutschland} \\
\textit{DB InfraGO AG}\\
Berlin, Germany \\
tobias.klockau@deutschebahn.com}
\and
\IEEEauthorblockN{Patrick Naumann}
\IEEEauthorblockA{\textit{Digitale Schiene Deutschland} \\
\textit{DB InfraGO AG}\\
Berlin, Germany \\
patrick.naumann@deutschebahn.com}
\and
\IEEEauthorblockN{Philipp Neumaier}
\IEEEauthorblockA{\textit{Digitale Schiene Deutschland} \\
\textit{DB InfraGO AG}\\
Berlin, Germany \\
philipp.neumaier@deutschebahn.com}
\and
\IEEEauthorblockN{Martin Köppel}
\IEEEauthorblockA{\textit{Digitale Schiene Deutschland} \\
\textit{DB InfraGO AG}\\
Berlin, Germany \\
martin.koeppel@deutschebahn.com}
}

\maketitle

\begin{abstract}
The monitoring of the route and track environment plays an important role in automated driving. For example, it can be used as an assistance system for route monitoring in automation level Grade of Automation (GoA) 2, where the train driver is still on board. In fully automated, driverless driving at automation level GoA4, these systems finally take over environment monitoring completely independently. With the help of artificial intelligence (AI), they react automatically to risks and dangerous events on the route. To train such AI algorithms, large amounts of training data are required, which must meet high-quality standards due to their safety relevance. In this publication we present an automatic method for assuring the quality of training data, significantly reducing the manual workload and accelerating the development of these systems. We propose an open-source tool designed to detect nine common errors found in multi-sensor datasets for railway vehicles. To evaluate the performance of the framework, all detected errors were manually validated. Six issue detection methods achieved 100\% precision, while three additional methods reached precision rates 96\% and 97\%.
\end{abstract}

\begin{IEEEkeywords}
quality checks, annotation, data sets, annotation errors
\end{IEEEkeywords}

\section{Introduction}

Automated trains, ranging from Grade of Automation (GoA) 2 to GoA 4, will be equipped with a wide range of advanced technologies. A key element of this is environmental and obstacle detection, which will be handled primarily on board the train. These systems employ radar, lidar, cameras, ultrasound, and infrared imaging to replicate the role of the human eye at the front and sides of the train \cite{r21}.

\begin{figure}
    \centering
    \subfloat[2D Bounding Box\label{1a_1}]{%
        \includegraphics[width=0.49\linewidth]{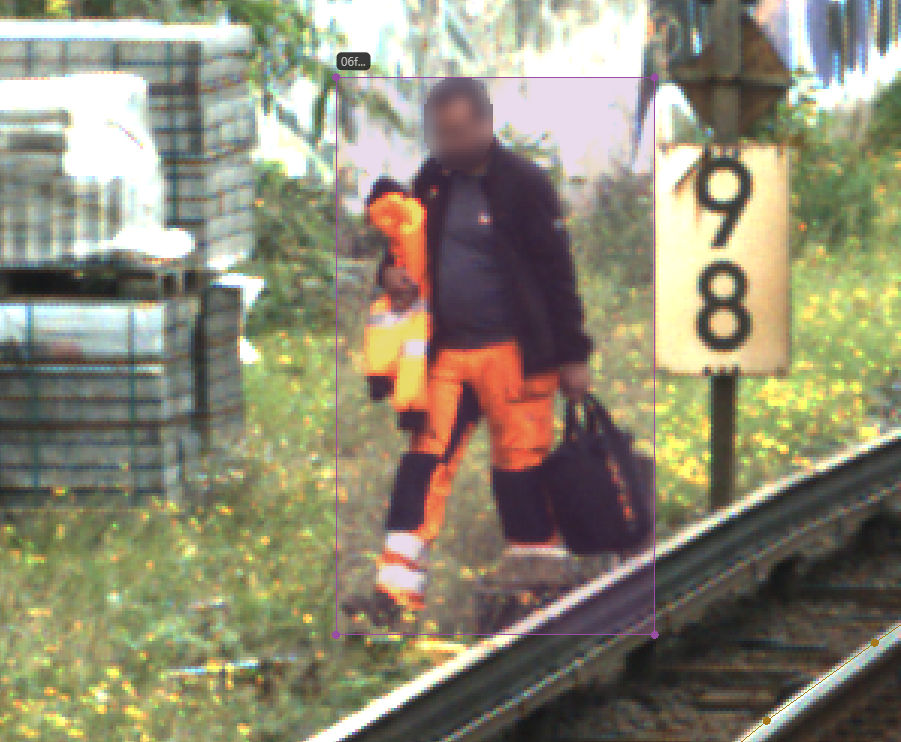}
        \hfill
    }
    \subfloat[Polyline\label{1b_1}]{%
        \includegraphics[width=0.49\linewidth]{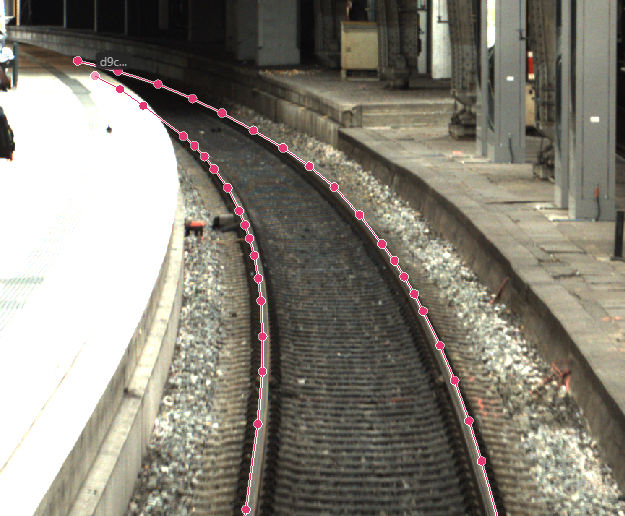}
    }
    \\
    \subfloat[Polygon\label{1c_1}]{%
        \includegraphics[width=0.49\linewidth]{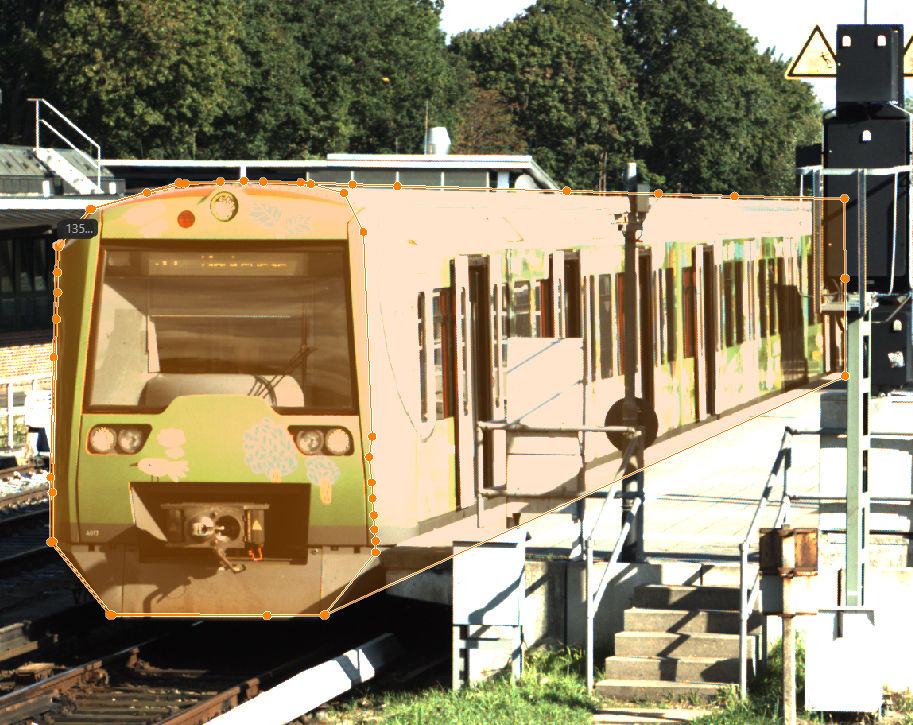}
        \hfill
    }
    \subfloat[3D Bounding Box\label{1d_1}]{%
        \includegraphics[width=0.49\linewidth]{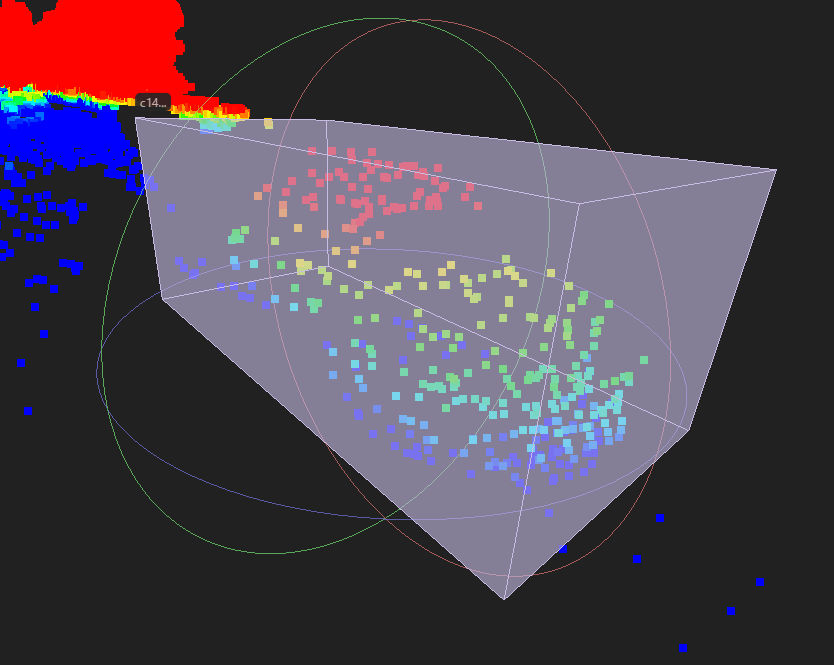}
    }
    \caption{Annotation types used in OSDaR23}
    \label{fig:annoation-type-examples}
\end{figure}

These environment perception systems monitor the surroundings and must detect obstacles in the track at an early stage. The technologies behind them are often  based on artificial intelligence (AI). Large amounts of data are required to train the AI for the task of monitoring the environment. This data is stored in a so-called “data factory” and processed into data sets for machine learning and AI in compliance with data protection regulations. The Data Factory - another DSD development project - provides a platform for storing, processing, simulating and annotating data \cite{a1}. This data consists of raw sensor data from cameras, several lidar and radar systems on the one hand and the associated manual annotations on the other. 
Annotations in sensor datasets, specifically for environmental detection, are markings in raw data for training, testing, and validating AI methods. For 2D sensors such as cameras, these annotations often include 2D bounding boxes [Fig. \ref{fig:annoation-type-examples}(a)], polylines [Fig. \ref{fig:annoation-type-examples}(b)], and polygons [Fig. \ref{fig:annoation-type-examples}(c)] to mark objects in images. For 3D sensors like lidar, point clouds are annotated by drawing 3D bounding boxes [Fig. \ref{fig:annoation-type-examples}(d)] around objects or marking specific points. The properties of individual annotations are represented by additional attributes, which consist of a label and a corresponding value.

\begin{figure*}
    \centering
    \subfloat[AnnotationAboveHorizon\label{1a}]{%
        \includegraphics[width=0.30\linewidth]{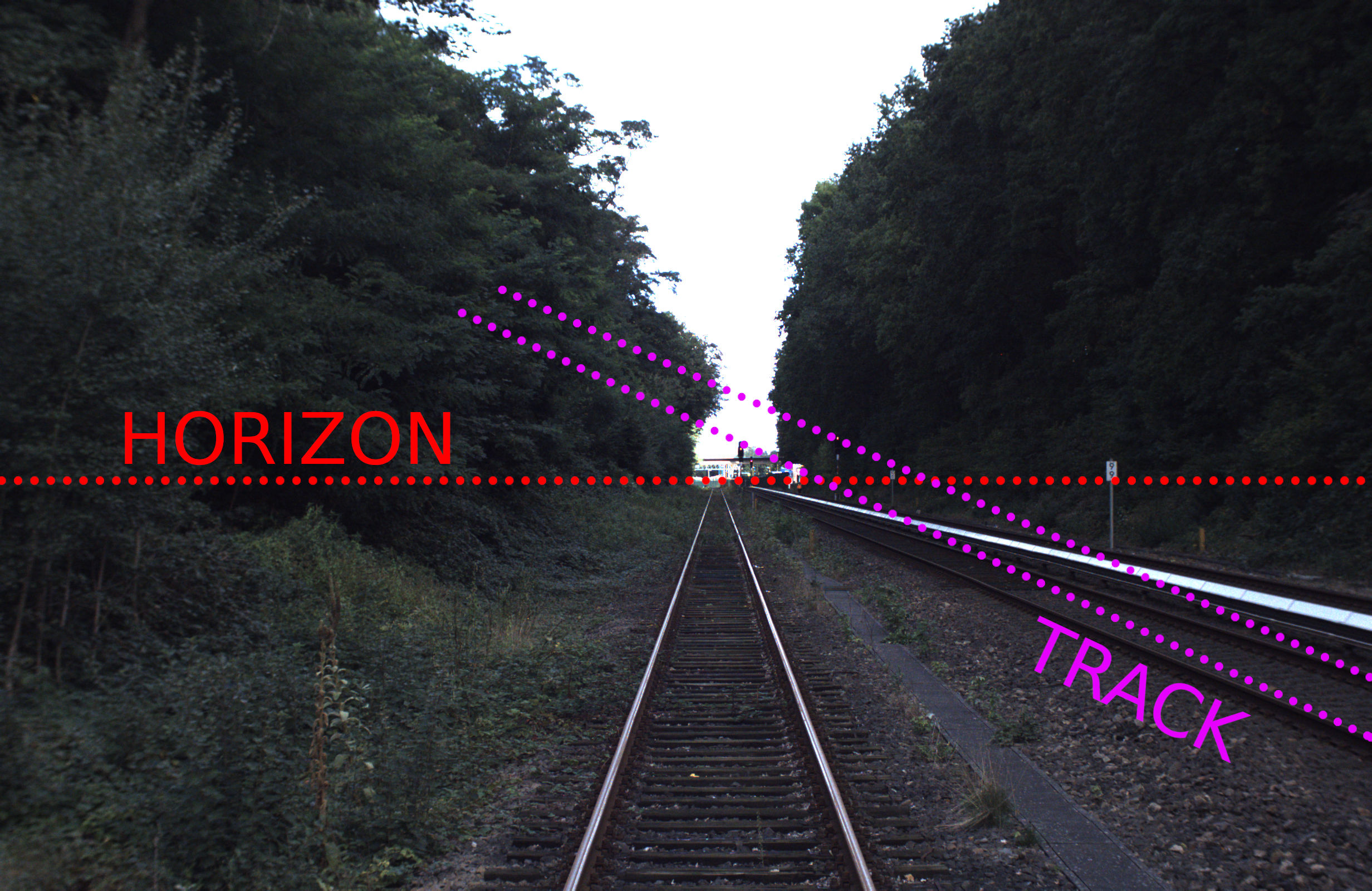}
        \hfill
    }
    \subfloat[DimensionInvalid\label{1b}]{%
        \includegraphics[width=0.20\linewidth]{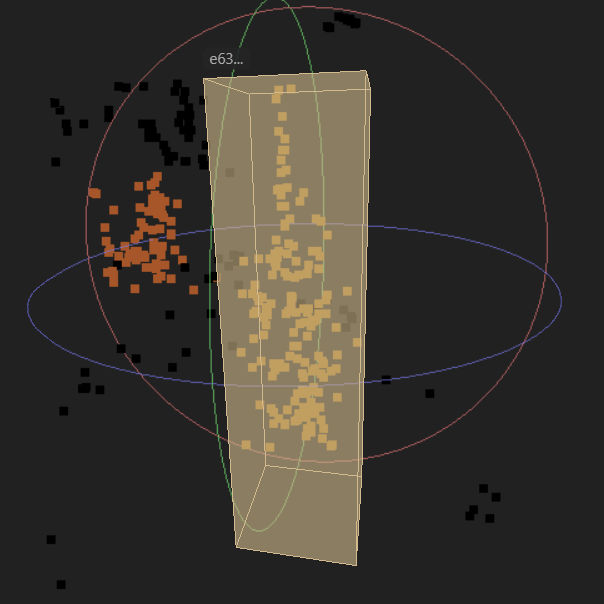}
        \hfill
    }
    \subfloat[InconsistentAttributeScope\label{1c}]{%
        \includegraphics[width=0.45\linewidth]{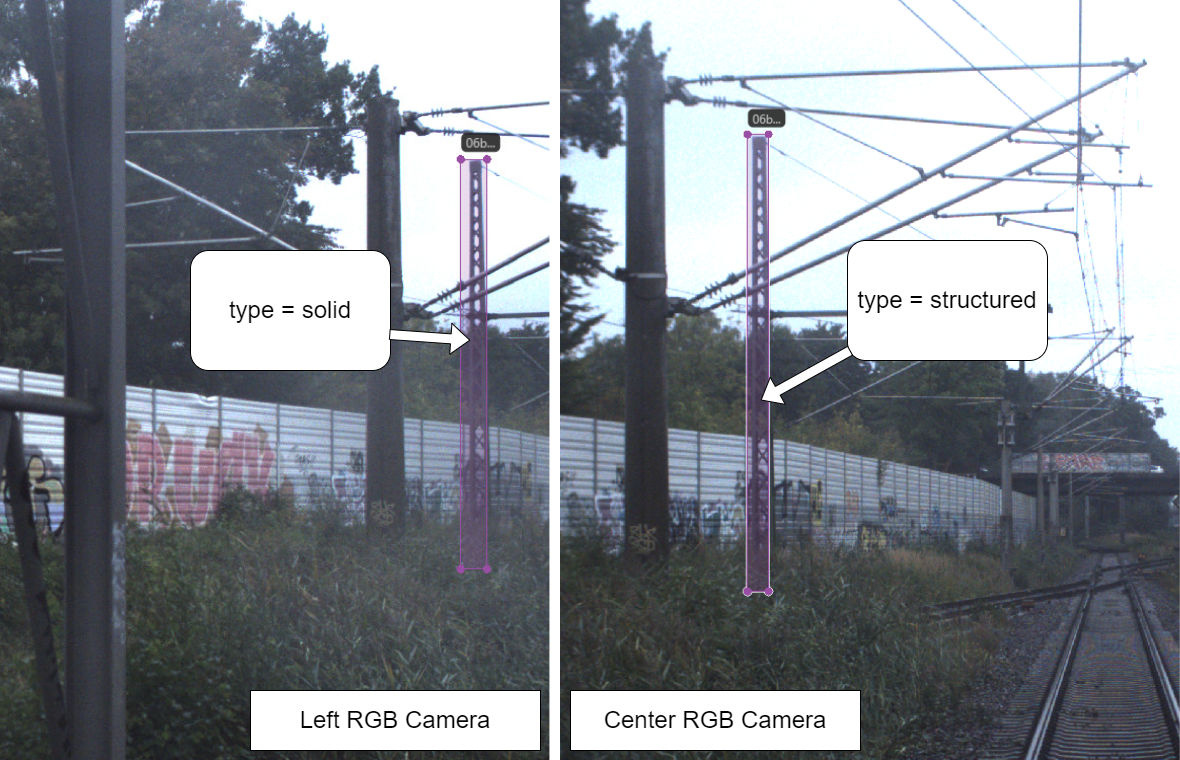}
        \hfill
    }
    \\
    \subfloat[MissingAttribute\label{1d}]{%
        \includegraphics[width=0.45\linewidth]{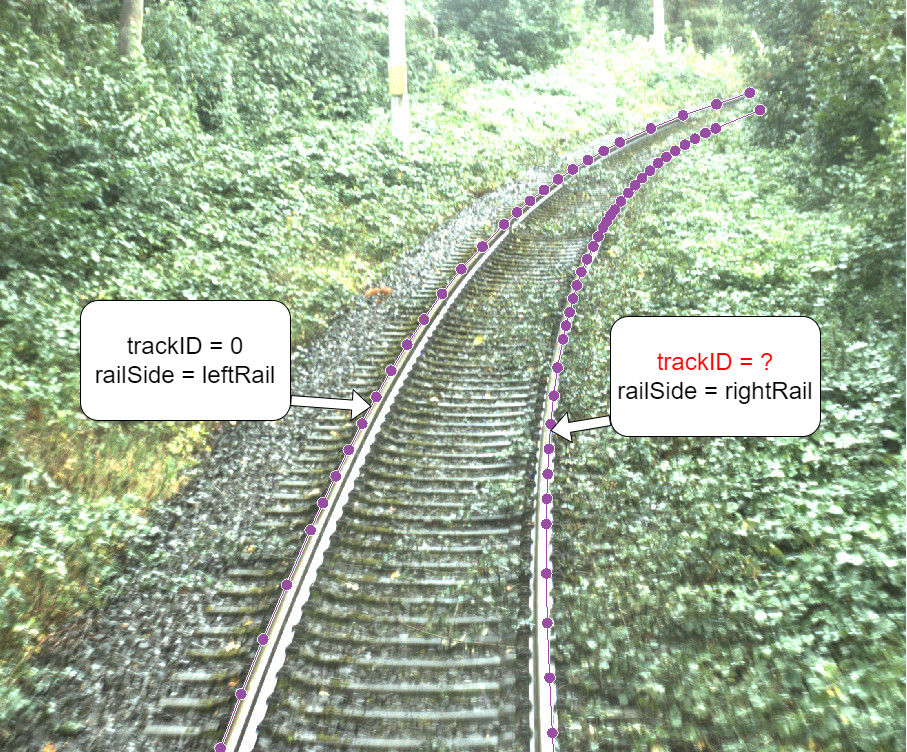}
        \hfill
    }
    \subfloat[UnexpectedAttribute\label{1e}]{%
        \includegraphics[width=0.45\linewidth]{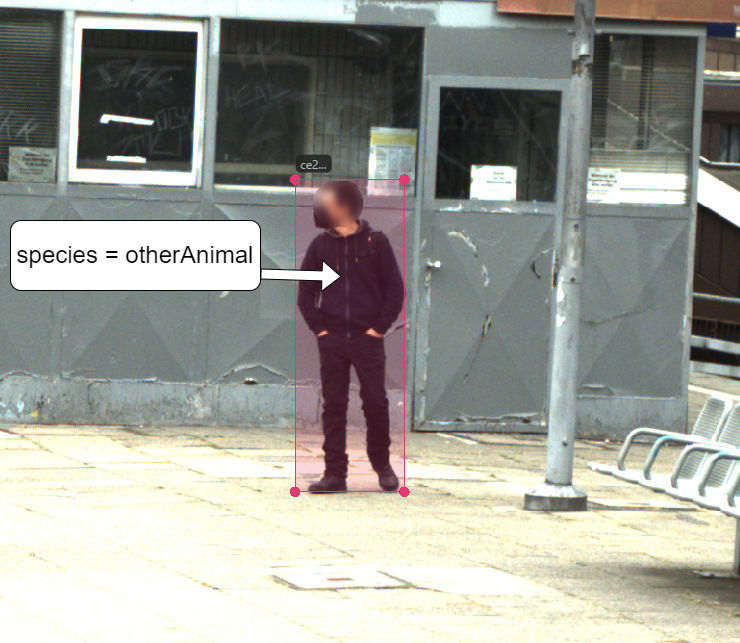}
        %\hfill
    }
    \caption{Issue Types covered by the automatic checks}
    \label{fig:error-overview_1}
\end{figure*}\textbf{}

\begin{figure*}
    \centering
    \subfloat[MissingEgoTrack\label{1f}]{%
        \includegraphics[width=0.49\linewidth]{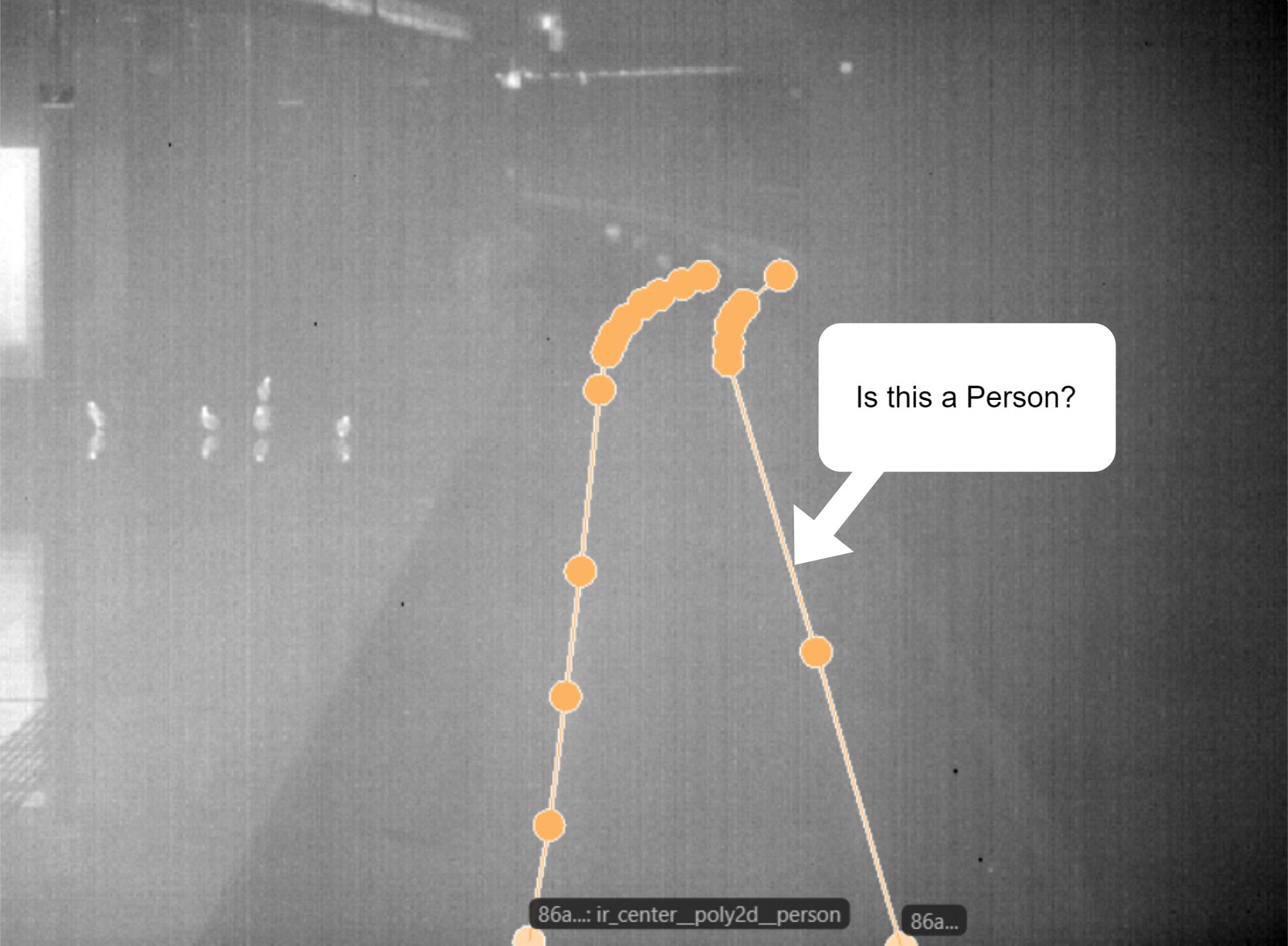}
        \hfill
    }
    \subfloat[RailSideCount\label{1g}]{%
        \includegraphics[width=0.49\linewidth]{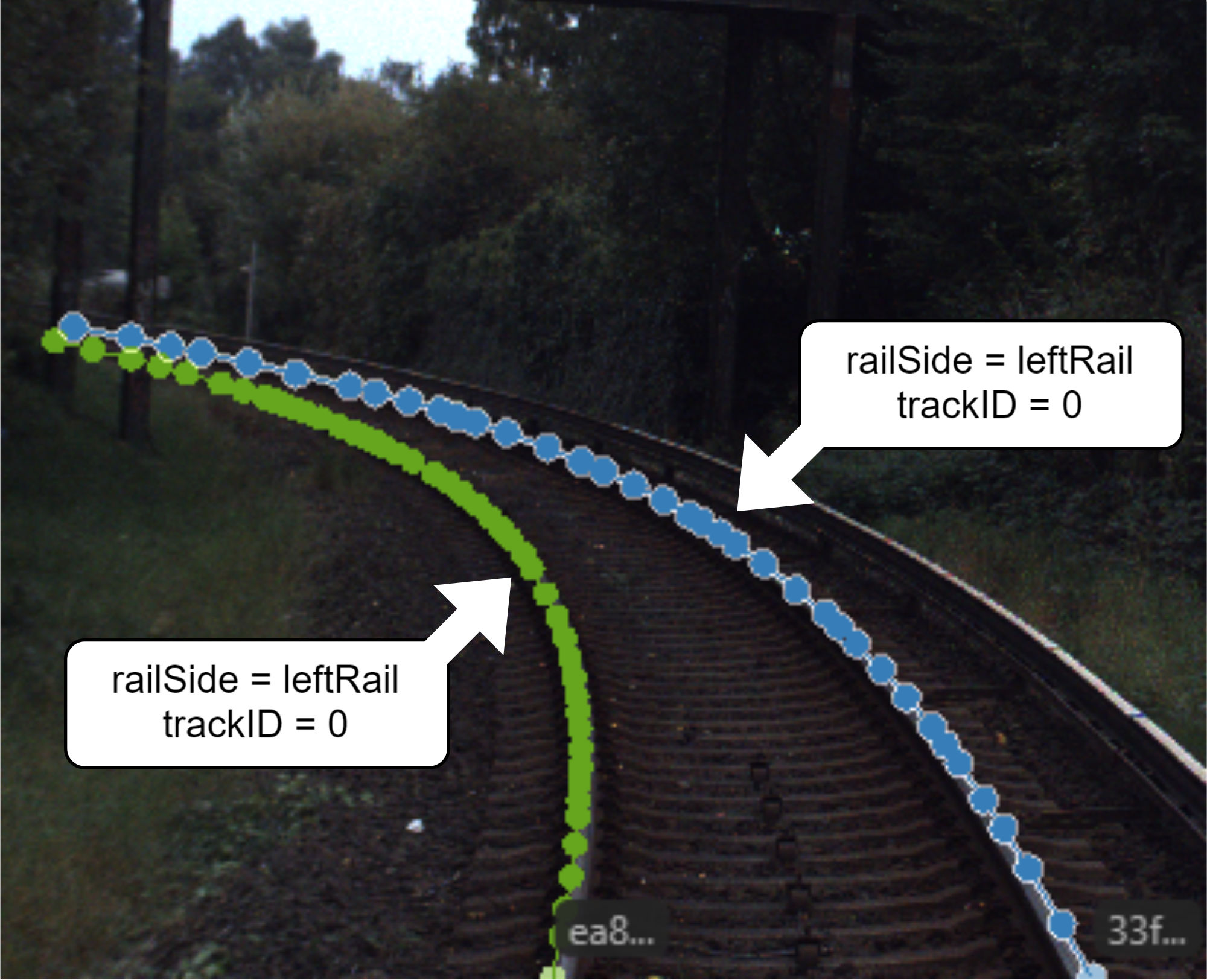}
        \hfill
    }
    \\
    \subfloat[RailSideOrder\label{1h}]{%
        \includegraphics[width=0.49\linewidth]{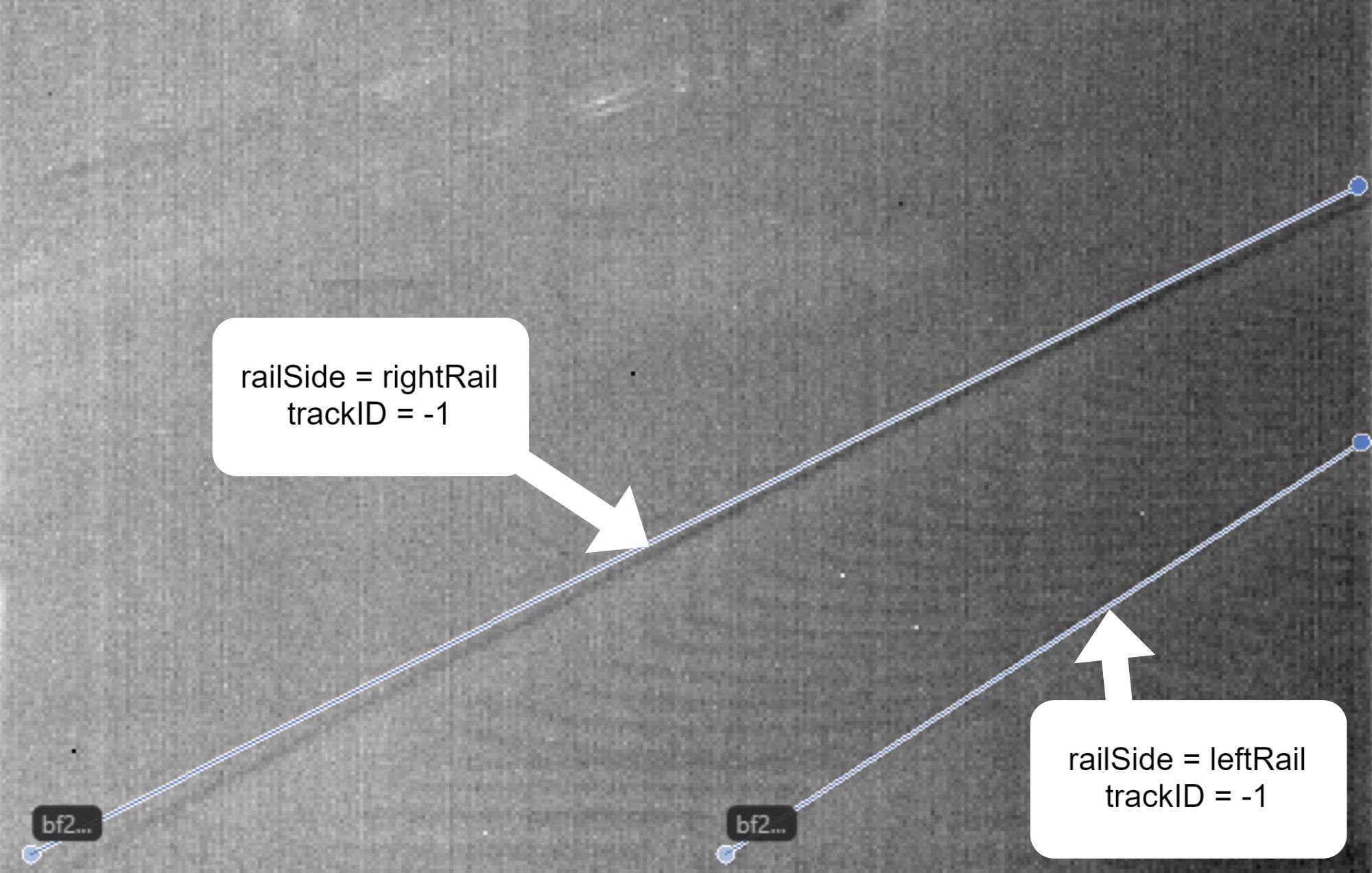}
        \hfill
    }
    \subfloat[TransitionIdenticalStartAndEnd\label{1i}]{%
        \includegraphics[width=0.49\linewidth]{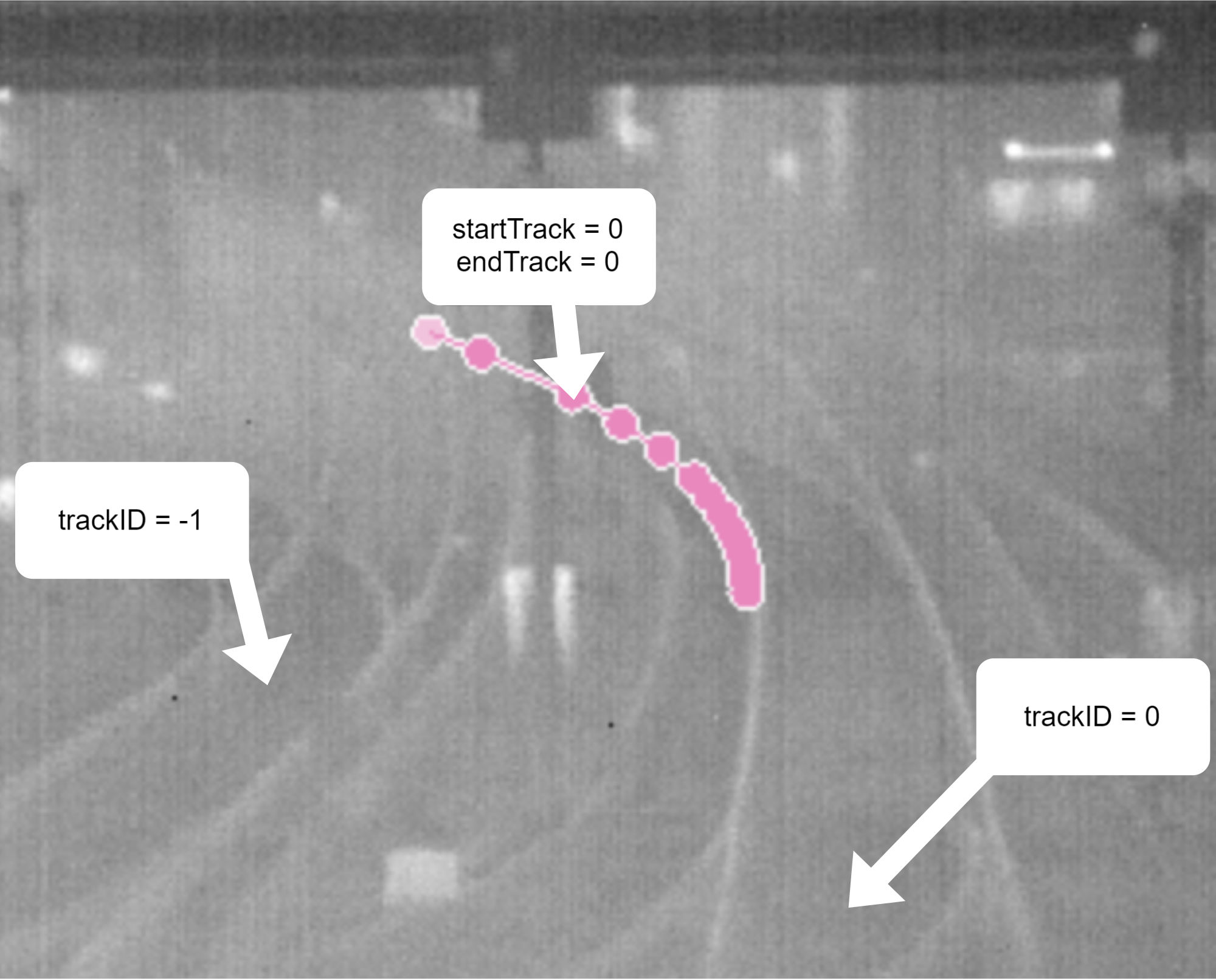}
        \hfill
    }
    \caption{Issue types covered by the automatic checks}
    \label{fig:error-overview_2}
\end{figure*}

Since environmental perception in rail transport is of safety-critical relevance, the annotations must meet a very high-quality standard. A thorough quality check is therefore required, which today is largely done manually. However, due to the constantly increasing amounts of data, it is becoming necessary to increasingly rely on automated error detection methods to enable faster data processing. The software presented in this publication enables a fast detection of some typical errors that are hard to find for humans. The software was tested on the publicly available OSDaR23 dataset \cite{a6}, \cite{a6_2}.

\pgfplotstableread[row sep=\\,col sep=&]{
    name & precision & remaining \\
    AnnotationAboveHorizon & 96 & 4 \\
    DimensionInvalid & 97 & 3 \\
    %EmptySensor & 24 & 76 \\
    InconsistentAttributeScope & 97 & 3 \\
    %InvalidAttributeType & 0 & 100 \\
    %InvalidAttributeValue & 0 & 100 \\
    MissingAttribute & 100 & 0 \\
    MissingEgoTrack & 100 & 0 \\
    RailSideCount & 100 & 0 \\
    RailSideOrder & 100 & 0 \\
    TransitionIdenticalStartAndEnd & 100 & 0 \\
    UnexpectedAttribute & 100 & 0 \\
    %UnknownObjectType & 0 & 100 \\
}\precisiondata
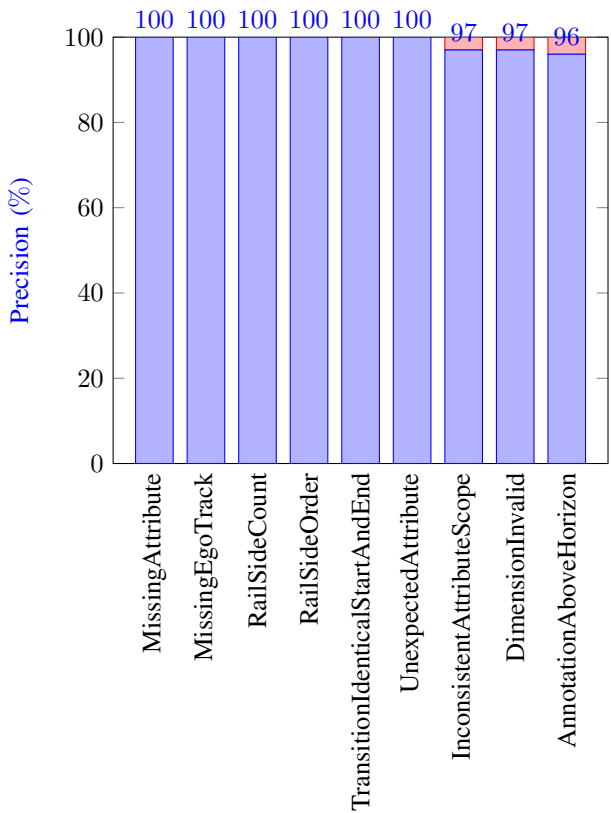
\begin{figure}
    \centering
    \begin{tikzpicture}
        \begin{axis}[
            ybar stacked,
            width=0.45\textwidth,
            height=0.4\textwidth,
            scaled ticks=false,
            tick label style={/pgf/number format/fixed},
            symbolic x coords={
                MissingAttribute,
                MissingEgoTrack,
                RailSideCount,
                RailSideOrder,
                TransitionIdenticalStartAndEnd,
                UnexpectedAttribute,
                InconsistentAttributeScope,
                DimensionInvalid,
                AnnotationAboveHorizon,
                %EmptySensor,
            },
            xtick=data,
            xticklabel style={rotate=90},
            bar width=0.5cm,
            ymin=0,
            ymax=100,
            ylabel={Precision ($\%$)},
            ylabel style={blue}
        ]
            \addplot+[nodes near coords] table[x=name,y=precision]{\precisiondata};
            \addplot table[x=name,y=remaining]{\precisiondata};
            %\legend{precision, $1-\text{precision}$}
        \end{axis}
    \end{tikzpicture}
    \caption[Observed precision of all issue types]{
        This chart shows the observed precision of each implemented issue type detector in percent.
    }
    \label{fig:precision-overview}
\end{figure}

The development of the quality checking tool has been finalized. Accordingly, the source code is made available to the research community and industry stakeholders to support further advancements in environmental monitoring systems.

\section{State-of-the-Art}

In manual inspection, all annotations are checked for issues, which is associated with a high workload. To reduce this effort, a representative sample of the data can be checked, but this can lead to a significant misjudgment of quality if the chosen sample is not ideal.

Certain types of errors can already be detected automatically \cite{a2}. The long-term objective is to enable automated quality checks to identify the vast majority of errors.
By establishing rules for detecting specific errors, the entire dataset can be effectively checked for these errors \cite{a3}. The necessary human effort for the inspection can thus be significantly reduced, as long as the accuracy of the detection methods has been validated. The automatically detected errors can then serve as a basis for the typically manual error correction \cite{a4}.
Building on automated error detection, it is possible to automatically correct errors under certain conditions. One simple approach is to delete all faulty annotations so that only correct annotations remain, at the cost of completeness \cite{a5}. Alternatively, annotations can be automatically modified. However, it is important to verify the correctness of the automatic error correction to ensure that no additional errors are introduced into the dataset.

In 2023, a multi-sensor dataset (OSDaR23) to develop fully automated driving methods for railway vehicles was published \cite{a6}, \cite{a7}, \cite{a8}. 
Furthermore, an annotation format that conforms with the ASAM openLabel Standard was proposed.
This dataset was created in cooperation between the German Center for Railway Research at the Federal Railway Authority (DZSF), DB InfraGO AG, and FusionSystems GmbH. The dataset contains 45 sequences of synchronized multi-sensor data. Three RGB cameras with medium resolution and three with high resolution, three infrared cameras, one radar, and six combined lidar sensors were used. The dataset contains a total of 204091 annotations. The requirements for the annotations are publicly available as a labeling guide \cite{a9}. To test our proposed automated error detection methods, the available OSDaR23 dataset was used. The OSDaR23 dataset had undergone a manual quality check iteration, meaning it was quality checked by humans.  We tested the proposed framework on the final release of the dataset and used the fully automated error detection methods for quality assurance. 

\section{Overview}

In the proposed framework, nine detectors were developed for the identification of annotation errors. Five of these detectors are specifically designed for the railway domain and, to the best knowledge of the authors, are introduced for the first time. The remaining four detectors are more general in nature and can be applied across multiple domains.

The development process is outlined in the following:
In an initial step, frequently occurring error types were systematically identified. Based on this analysis, a set of detection rules was formulated and subsequently implemented as software algorithms. The implementation was carried out in Python. Then the proposed framework was tested on the OSDaR23 dataset  \cite{a6}, \cite{a7}, \cite{a8}. Finally, the discovered errors were checked by a human, to determine the precision of the quality checks.

\section{Problematic Annotations and Detection}
In this section the issue types which can be detected with the proposed framework are described.
Figures \ref{fig:error-overview_1} and \ref{fig:error-overview_2} show examples for these errors.
%In Figure \ref{fig:error-overview_1} examples for (a) AnnotationAboveHorizon, (b) DimensionInvalid, (c) InconsistentAttributeScope,  (d) MissingAttribute and (e) UnexpectedAttribute are presented.
%In Figure  \ref{fig:error-overview_2} examples for (a) MissingEgoTrack, (b) RailSideCount,  RailSideOrder (c) and TransitionIdenticalStartAndEnd (d) are shown.

The error types AnnotationAboveHorizon, MissingEgoTrack, RailSideCount, RailSideOrder, and TransitionIdenticalStartAndEnd are specific to the railway domain. In contrast, the remaining error types are more generic and can also be applied to validate datasets from other domains.
The detailed description of the error types is given in the following:
%All of these issues can be uncovered with the proposed software.

\begin{itemize}
    \item AnnotationAboveHorizon [cf. Fig. \ref{fig:error-overview_1}(a)]: If a track is drawn too far and extends beyond the horizon, it is classified as the error type AnnotationAboveHorizon. In the example, a track extends into the sky. This issue type is checked for all camera sensors.
    \item DimensionInvalid [cf. Fig. \ref{fig:error-overview_1}(b)]: 
    If an object exceeds a specified size, it is classified as the error type DimensionInvalid. It is checked if an object is too big or too small compared to its expected size. In the shown example, a person is larger than 3m. This type of error is checked for the merged 3D point cloud, i.e. the lidar sensors.
    \item InconsistentAttributeScope [cf. Fig. \ref{fig:error-overview_1}(c)]: 
If an attribute that is expected to remain constant across images or sensors changes, it is classified as the error type InconsistentAttributeScope. In the example, the “Type” of a “Catenary Pole” is labeled as “structured” in one camera view and as “solid” in another. This type of error is checked for all sensors.
    \item MissingAttribute [cf. Fig. \ref{fig:error-overview_1}(d)]:
The MissingAttribute error type occurs when an annotation lacks an expected attribute. In the example, the “Track” annotation is missing the “TrackID” attribute. This type of error is checked for all sensors.
    \item UnexpectedAttribute [cf. Fig. \ref{fig:error-overview_1}(e)]: In the UnexpectedAttribute error type, an annotation has an unexpected attribute for its object class. For example, the attribute "Species" was set for a person, which actually belongs to the object class "Animal" [cf. Fig. \ref{fig:error-overview_1}(e)]. This type of error is checked for all sensor.
    \item MissingEgoTrack [cf. Fig. \ref{fig:error-overview_2}(a)]:
The error type MissingEgoTrack occurs when the track on which the train is traveling is not annotated as the “Ego-Track” in the center camera view. Additionally, an “Ego-Track” annotation must be present in the merged lidar point cloud. In the example, the track is incorrectly labeled as a “Person.” This error type is checked for both the center camera and the merged point cloud.
    \item RailSideCount [cf. Fig. \ref{fig:error-overview_2}(b)]: The RailSideCount error type occurs when a track has more than one left or right rail. Normally, a track consists of a single left and a single right rail. In cases of occlusion, only one left or right rail may be visible. This error is checked across all camera sensors.
    \item RailSideOrder [cf. Fig. \ref{fig:error-overview_2}(c)]: The RailSideOrder error type occurs when the left and right rails on a track are reversed. In the example, the left rail appears on the right side, and the right rail appears on the left. This error is checked across all camera sensors.
    \item TransitionIdenticalStartAndEnd [cf. Fig. \ref{fig:error-overview_2}(d)]: The TransitionIdenticalStartAndEnd error type occurs when a transition track, intended to connect two different tracks, starts and ends on the same track. In the example shown, the transition begins on trackID 0 (the current track) and is supposed to end on the neighboring track (trackID \mbox{-1}), but it erroneously ends on trackID 0. This error is checked across all sensors.
\end{itemize}

%\begin{table}[htbp]
%\caption{Table Type Styles}
%\begin{center}
%\begin{tabular}{|c|c|c|c|}
%\hline
%\textbf{Table}&\multicolumn{3}{|c|}{\textbf{Table Column Head}} \\
%\cline{2-4} 
%\textbf{Head} & \textbf{\textit{Table column subhead}}& \textbf{\textit{Subhead}}& \textbf{\textit{Subhead}} \\
%\hline
%copy& More table copy$^{\mathrm{a}}$& &  \\
%\hline
%\multicolumn{4}{l}{$^{\mathrm{a}}$Sample of a Table footnote.}
%\end{tabular}
%\label{tab1}
%\end{center}
%\end{table}

% TODO: Stichpunkte ausformulieren
\section{Open Source Release}
 %   \begin{itemize}
 %       \item The validation functionality is implemented in the open-source \href{https://github.com/DSD-DBS/raillabel-providerkit}{raillabel-providerkit} project
  %      \item accessible as a Python library via pip
  %      \item based on the \href{https://github.com/DSD-DBS/raillabel}{raillabel} library, which provides a way to parse the OSDaR23 data in Python
  %     \item The uploaded code can differ from the code used for this paper due to improvements / refactoring
 % \end{itemize}

The proposed software for checking annotations is provided as open-source code, i.e. the  \href{https://github.com/DSD-DBS/raillabel-providerkit}{RailLabel-providerkit} \cite{a10} . The Python Library is accessable via pip, which makes it easy to install and to use. The RailLabel-providerkit builds upon the \href{https://github.com/DSD-DBS/raillabel}{RailLabel} library \cite{a11}, which offers a straightforward interface for accessing the OSDaR23 dataset in Python. The validation process is limited to the analysis of the annotation files (JSON) and does not rely on the associated sensor data. As a result, the error detection can be performed independently of the raw sensor recordings. Thereby the application of the tool is simplified.

\section{Evaluation}

The developed framework was evaluated using the publicly available OSDaR23 dataset  \cite{a6}, \cite{a6_2}. This dataset had previously undergone manual quality inspection. Therefore, any errors identified by our method represent issues that were not detected by human reviewers beforehand.

To evaluate the precision of the developed error detection methods, all identified errors were manually reviewed. This process enabled us to determine whether each case represented a true error or a false positive generated by the software. Figure \ref{fig:precision-overview} presents, for each error type, the percentage of automatically detected errors that were confirmed as genuine during manual inspection.

For six error types, the developed detection methods achieve a precision of 100\%, indicating that all automatically detected annotation errors were confirmed as actual errors. For three additional error types, the precision was between 96\% and 97\% (cf. Fig. \ref{fig:precision-overview}). In these cases, the software produced 3–4\% false positives that were subsequently dismissed during manual review.

A detailed analysis revealed that, for the DimensionInvalid error type, the detection thresholds defined in the rules did not align with certain objects. For instance, a construction vehicle with a fully extended excavator arm exceeded the expected z-dimension. This highlights a trade-off between enforcing strict 3D bounding box size limits and the risk of missing actual errors. Consequently, a small number of false positives is considered acceptable.
Additionally, inaccuracies in motion estimation can result in enlarged objects within merged point clouds, occasionally causing the size thresholds to be exceeded.
For the AnnotationAboveHorizon error type, a trade-off was made between the flat-earth assumption, which can lead to false positives, and the risk of overlooking actual errors. The flat-earth assumption was used to determine the height of the horizon in the images, based on the calibration of the actual sensor setup.
For the InconsistentAttributeScope issue type, the manually reviewed 3\% of false positives revealed edge cases that had not been addressed in the original requirements. These findings show the need to refine the annotation requirements in future work.

To determine the overall proportion of incorrect annotations, all potential sources of issues were aggregated. This included all annotation types, i.e. 2D bounding boxes (2DBB), 3D bounding boxes (3DBB), polylines, and polygons along with their associated attributes. In total, these accounted for 1,651,208 annotation elements. The proposed methods detected 35,931 errors across this dataset, indicating that 2.18\% of the annotations and their attributes contain errors.

\section{Conclusion and Future Work}

We have presented a software tool capable of identifying various types of errors in annotated multi-sensor datasets. In total, nine error types can be detected. Five of the identified error types are specific to the railway environment, while the remaining four represent domain-independent issues that may occur across various application contexts. To assess the reliability of the system, all detected errors were manually validated to distinguish true errors from false positives. Six of the error detection methods achieved a precision of 100\%, while three others demonstrated precision rates between 96\% and 97\%.

With this work, DB InfraGO AG has made an important contribution and sent a strong signal to the industry. The software is publicly available to interested parties and partners, serving as a foundation for future developments in both industry and research.

\section*{Acknowledgement}

We would like to thank Seo-Young Ham and Pavel Klasek  from the DZSF for their support during the work.
For visualizing the proposed data and its annotations, the WebLabel player provided by VICOMTECH was utilized \cite{a12}.

\end{document}